\documentclass[letterpaper, 10 pt, conference]{ieeeconf}  

\IEEEoverridecommandlockouts                              

\usepackage{graphics} 
\usepackage{graphicx}
\usepackage[style=ieee,backend=biber]{biblatex}
\usepackage{amsmath} 
\usepackage{amssymb}  
\usepackage{booktabs}
\usepackage{caption}
\usepackage{multirow}
\usepackage{pifont}
\usepackage[table]{xcolor}
\usepackage{cleveref}
\usepackage{dsfont}
\usepackage{listings}
\usepackage{longtable}
\let\labelindent\relax
\usepackage{enumitem}
\usepackage{float}\usepackage{lipsum}
\title{\LARGE \bf
Domain-Conditioned Scene Graphs for State-Grounded Task Planning 
}

\author{Jonas Herzog$^{1}$, Jiangpin Liu$^{1}$, Yue Wang$^{1}$
\thanks{This work was supported in part by the Joint Funds of the National Natural Science Foundation of China under Grant No. U24A20128, and in part by Zhejiang Provincial Natural Science Foundation of China under Grant No. LD25F030001.}
\thanks{$^{1}$Zhejiang University
}%
}
\begin{document}

\maketitle
\thispagestyle{empty}
\pagestyle{empty}

\begin{abstract}

Recent robotic task planning frameworks have integrated large multimodal models (LMMs) such as GPT-4o.
To address grounding issues of such models, it has been suggested to split the pipeline into perceptional state grounding and subsequent state-based planning.
As we show in this work, the state grounding ability of LMM-based approaches is still limited by weaknesses in granular, structured, domain-specific scene understanding. 
To address this shortcoming, we develop a more structured state grounding framework that features a domain-conditioned scene graph as its scene representation.
We show that such representation is actionable in nature as it is directly mappable to a symbolic state in planning languages such as the Planning Domain Definition Language (PDDL).
We provide an instantiation of our state grounding framework where the domain-conditioned scene graph generation is implemented with a lightweight vision-language approach that classifies domain-specific predicates on top of domain-relevant object detections.
Evaluated across three domains, our approach achieves significantly higher state grounding accuracy and task planning success rates compared to LMM-based approaches.
\url{https://github.com/Vision-Kek/DC-SGG}

\end{abstract}


\section{INTRODUCTION}
Task planning in a real environment relies on two core capabilities: (a) reasoning to find an action plan that fulfills the goal, and (b) scene understanding to accurately recognize the state of the environment \cite{sayplan}.
Traditionally, these capabilities had to be learned through in-domain training, which resulted in models that could only perform well within specific tasks, objects, or environments. 
Large Language Models (LLMs) with their strong generalization offer the potential to overcome these limitations and have therefore gained considerable attention as general task planners \cite{saycan}. 
Their extension to multimodality (LMMs) furthermore promised to enable joint reasoning over scene observation and instruction.

\begin{figure}[t]
    \centering
    \includegraphics[width=\linewidth]{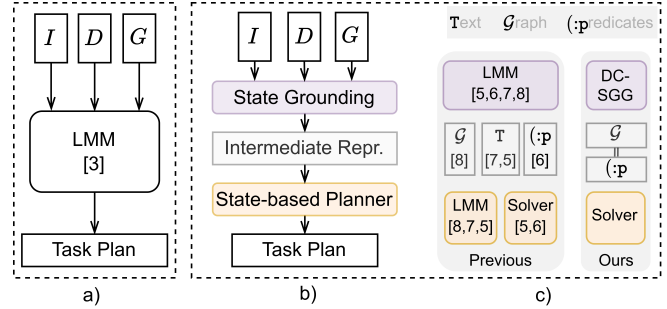}
    \caption{a) Direct task planning from $I$mage, $D$omain specification and $G$oal instruction \cite{replanvlm}. b) Enhancing grounding through an Intermediate Representation \cite{vilain,neuroground,vila,verigraph}. c) We perform domain-conditioned scene graph generation (DC-SGG) to extract a representation in a structure that is directly mappable to plannable PDDL predicates.}
    \label{fig:intro}
\end{figure}

The straightforward approach \cite{replanvlm} would be to prompt the LMM with the observation and instruction inputs and let it directly output a task plan (\Cref{fig:intro}a).
Studies \cite{sayplan,groundeddecoding} pointed out that grounding them in reality is the main challenge.
One point of grounding is to enforce adherence to robot- and domain-specific constraints such as the available skills, affordances and rules \cite{sayplan}.
The other side of grounding is they must identify the domain-relevant state of the environment, i.e. what are the relevant objects in the scene and what is their current configuration \cite{neuroground}.
Recent works \cite{vilain,neuroground,vila,verigraph,grid} have therefore suggested to build on an intermediate state representation (\Cref{fig:intro}b).

Among them, Shirai et al. \cite{vilain} provide structure for constraints and state representation by utilizing Planning Domain Definition Language (PDDL) \cite{pddl} which is used by classic symbolic planning systems.
A PDDL \emph{domain} specifies robot-specific skills and domain-specific \emph{lifted} predicates, such as ``carries'' or ``in''.
The objective is then to transform input scene observation and task instruction into a PDDL \emph{problem}, which \emph{grounds} the predicates into initial state, e.g. ``carries(robot,cucumber)'', and goal state, e.g. ``in(cucumber,bowl)''.
A symbolic planner can find an optimal plan that transforms initial state to goal state, even for large state spaces, a problem LLMs struggled at \cite{llmp}.

Outsourcing (a) ``finding an action plan'' to a symbolic planner, shifts the problem focus to (b) ``recognition of the environment state''.
Previous works generate initial and goal state either by letting an LLM reason on top of detections of vision models \cite{vilain}, or, more recently, by directly processing scene observation, domain specification and task instruction in a single multimodal such as GPT-4o \cite{vila,neuroground}.
However, as we show in this paper, LMM-based parsing of the scene into the correct initial state is currently the major bottleneck for task planning success.
We therefore aim for a more accurate scene parser that yields a plannable representation.
We formulate the problem as domain-conditioned scene graph generation, where the task is to predict a set of task-relevant objects (nodes) and domain-relevant relationships (edges).
This opens the door for leveraging ideas from scene graph research while maintaining a representation structure that complies with PDDL, and is hence, in contrast to general purpose scene graphs, directly actionable.
We propose an LMM-free baseline that builds a simple relationship classifier on top of described \cite{describedod} object detection.
By comparing (\Cref{fig:intro}c) our method with the different approaches for state grounding and state-based planning, we expose current limitations of LMM-based methods and the challenges to solve.
Our contributions are therefore:
\begin{enumerate}[leftmargin=14pt]
\item We propose a state grounding framework that parses the scene into a PDDL-aligned domain-conditioned scene graph, improving state grounding accuracy and task planning success over previous frameworks.
\item Through comparison with (i) LLMs reasoning over perception data,  (ii) LMMs-As-Planners, (iii) LMMs as predicate estimators, and (iv) open-vocabulary scene graph generators, we highlight the shortcomings of (i-ii) as well as possibilities and remaining challenges of (iii,iv).
\end{enumerate}

\section{RELATED WORK}
\subsection{Task Planning from Observed State}
Planning on top of a visually estimated state is a classic pipeline approach.
Structuring the state representation in an object-centric and symbolic form \cite{rpn,geometricsymbolic,sornet} further allows to find plans through logic search on the symbolic level.
The symbolic state can be visually estimated through learning predicate classifiers \cite{groundingpredicatesthroughactions,kase,sornet} grounded on object detections.
However, these methods do no not generalize beyond a closed set of predicates and objects or environments seen during train-time.
In contrast, due to their common world knowledge and reasoning abilities, large pre-trained models promise to overcome this limitation. 
Robotic task planning with large language models (LLMs) and vision language models (VLMs) has therefore been studied \cite{sayplan,saycan,vilain}, furthermore allowing the goal to be specified in natural language.
Operating on text modality, visual grounded task planning with LLMs needs VLMs to transmit abstracted visual information to the LLM.
This might lose task-relevant information and prohibits joint reasoning over vision and language modalities \cite{vila}.
Hence, multimodal L\emph{M}Ms\footnote{We use the term LMM instead of VLM to distinguish approaches operating in open-ended text-generative LLM mode (LMM) from vision-centered models like open-vocabulary object detectors (VLM).}
came into the focus of recent research \cite{replanvlm,vila,neuroground}, of which \cite{replanvlm} directly schedules actions, needing another LMM for error correction, \cite{vila} builds a language intermediate scene representation through chain-of-thought prompting and \cite{neuroground} explicitly prompts for initial state, goal state and action plan to be sequentially generated.
In other words, these approaches use generated text as an intermediate representation which is not strictly bound to a structure.

\subsection{PDDL Problem as State Representation}
Tasks that should follow more structured, well-defined and human-interpretable \cite{rpn} constraints can be specified in a formal language such as Planning Domain Definition Language (PDDL) \cite{pddl}. %
Kase et al. \cite{kase} learn to predict the PDDL initial state from observation, but require in-domain training, hence are limited to seen objects, predicates and actions.
Research on LLM-planning provided the state representation in PDDL  \cite{generalizedplanning}, or equipped the LLM with an algorithmic PDDL solver to achieve correct optimal plans \cite{llmp}.
The latter is similar to our motivation, we use a PDDL solver and follow the problem setting of \cite{vilain} that the domain file specifying the constraints is given and the problem file specifying the scene's initial and goal state is to be generated.

\subsection{Scene Graph as a Plannable Scene Representation}
Ray et al. \cite{tampin3dsg} perform Task and Motion Planning (TAMP) in given hierarchical  3D scene graphs designed for large scale environments. 
While they also use PDDL predicates, they need to infer them on top of the predefined graph structure, which restricts the set of inferrable predicates and thus also the range of solvable tasks.
Also within a TAMP framework, Zhu et al. \cite{geometricsymbolic} construct scene graphs at the geometric and symbolic level, allowing motion planning on the geometric and task planning on the symbolic scene graph.
Their basic idea to map a geometric scene graph to a symbolic scene graph is similar to ours, but the mapping is manually defined and the set of predicates is fixed.
ConceptGraphs \cite{conceptgraphs} incrementally build a 3D scene graph through fusing the predictions of vision models into a set of entities and annotating their properties and relations with the help of vision-language models.
The shortcoming of this approach is that the decomposition of the scene and generation of annotations is generic and not task- and domain-conditioned, hence structure and information may not align with the target task.
Among the LLM-based works \cite{sayplan,llmstate,grid,verigraph}, SayPlan \cite{sayplan} lets an LLM search and plan in a 3D scene graph, solving scalability issues.
Ni et al. \cite{grid} transform scene graphs to tokens to train an LLM-Encoder Action-Decoder network, showing superior performance over LLM-as-Planner approaches, but requiring training on scene- and robot-specific data.
VeriGraph \cite{verigraph} and \cite{sequentialmanipulationonsg} model actions as graph edit operations, whereof \cite{verigraph} uses LMMs for both scene graph generation and planning on the scene graph.
Similar to other LMM-based methods \cite{sayplan,replanvlm,vilain}, the open-ended generative nature of LLMs leads to incorrect propositions, requiring multiple iterations of error-corrective reprompting.

\vspace{4pt}
In this work, we bring the ideas of PDDL as state representation and scene graphs as scene representation together, introducing domain-conditioned scene graphs and exploring how such structure can be generated.

\begin{figure*}[htbp]
    \centering
    \includegraphics[width=0.85\linewidth]{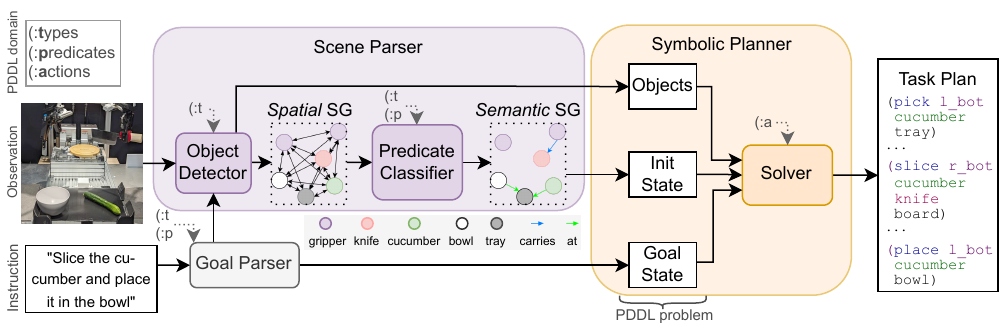}
    \caption{Overall system to generate task plan $P$ from PDDL domain specification $D$, image observation $I$, and instruction $G$. The scene representation generated by the scene and goal parser is conditioned on \textbf{t}ypes and \textbf{p}redicates of the domain, producing a structure that complies with the PDDL domain specification and thus enables the use of classic symbolic planners to find an optimal solution. 
    }
    \label{fig:overall}
\end{figure*}

\section{State-Grounded Task Planning}
\subsection{Problem Formulation}
Given domain specification $D$, goal or task instruction $G$ and visual scene observation $I$, the system $\Phi$ should output a task plan $P$ :
\begin{equation}
    P=\Phi(D,G,I).
\end{equation}

The task plan is a sequence  \( P = [a_1(o_1), \dots, a_n(o_n)] \) of actions \( a_i(o_i) \), where $a_i \in D_{a}$ is a robot skill that is grounded on $o_i \subseteq \mathcal{O}$, where $\mathcal{O}$ are the task-relevant objects in the environment.

The robot must adhere to domain-specific constraints, which are given within domain specification $D$.
It provides the robot skill set $D_a$, where each contained lifted action $a_i \in D_a$ may operate on a number of object types $\in D_t$ and has preconditions and effects, of which both are a combination of lifted predicates $\in D_p$.
Skill set, domain-relevant object types $D_{t}$ and lifted predicates $D_{p}$ compose 
\begin{equation}
    D=(D_{t},D_{p},D_{a}).
\end{equation}
This 3-tupled domain specification aligns with the structure of a PDDL \emph{domain} file.

Methods that use an intermediate state representation solve the problem two-staged: $\Phi=f_{plan}\circ f_{state}$. 
State grounding $f_{state}$ generates grounded object set, initial and goal state as
$$\mathcal{O}, S_{init},S_{goal} = f_{state}(D,G,I).$$
This 3-tupled state representation aligns with the structure of a PDDL \emph{problem} file.

$S_{init},S_{goal}$ are plannable if they are composed of domain-specific grounded predicates:
\begin{equation}
S_{init}, S_{goal} \subseteq \{ p(o) \mid p \in D_p, \ o \in \mathcal{O}^k \},    
\label{eq:plannablecondition}
\end{equation}
where $p(o)$ is a grounded predicate over $k$ objects, e.g. unary $k=1$ for object properties and binary $k=2$ for object relationships.
The ``plannability'' through composition of $p \in D_p$ is because preconditions and effects of actions are composed of $p \in D_p$ as well, so that planner $f_{plan}$ can then operate on the symbolic level to find an action sequence $P$ whose cumulative effects transform $S_{init}$ to $S_{goal}$.

\subsection{Domain-Conditioned Scene Graph}
\label{sec:probsetdcg}
We show how scene graphs are directly plannable if they are domain-conditioned.
A conventional scene graph is commonly defined as $\mathcal{G} =(V,E)$,
where
\begin{itemize}
    \item an edge $e \in E$ forms a triplet $e_{ij}=(v_i,r,v_j)$
    \item connecting two vertices $v_i,v_j \in V$
    \item with relationship $r\in R$.
\end{itemize}
In this general form, $\mathcal{G}$ might contain task-irrelevant information while lacking task-relevant information, so that knowledge about the current environment state is incomplete, forbidding planning.
Now, critically, $\mathcal{G}$ becomes
\begin{itemize}
    \item \emph{domain-conditioned} through $V=\mathcal{O}$ and $R=D_p$,
    \item and hence \emph{plannable} through \Cref{eq:plannablecondition}.
\end{itemize}
We therefore identify a domain-conditioned scene graph as the optimal structure for our state-grounded task planning problem.

\section{METHOD}
Our overall system for generating task plan $P$ based on domain specification $D$, instruction $G$ and scene observation $I$ is illustrated in \Cref{fig:overall}.

$f_{state}$ from the problem definition needs to generate initial and goal state.
We identified that generating the correct initial state is the main weakness in previous frameworks. 
Therefore, in \Cref{subsec:sceneparser} we dedicate our attention to the design of a scene parser.
The goal parser is then covered by \Cref{subsec:goalgen}.
Unlike previous frameworks, we generate the goal state first, independent from the harder to estimate initial state, avoiding error propagation.
Given initial and goal states and benefiting from the structured plannable representation, we can follow \cite{vilain} to use an off-the-shelf solver \cite{fastdownward} for $f_{plan}$.

\subsection{Scene Parser}
\label{subsec:sceneparser}
Given the current observation and domain specification, the goal of the scene parser is to generate a plannable representation.
From the perception perspective, this is a scene representation, from the planning perspective it corresponds to initial state $S_{init}$.

We propose to realize scene parsing through generation of a domain-conditioned scene graph $\mathcal{G}=(V,E)$.
We show an implementation on top of an object detector supporting language queries.
First, a spatial scene graph is generated from the bounding box information.
Subsequently, the spatial scene graph is mapped to a semantic scene graph that matches the constraints from \Cref{sec:probsetdcg}.

\subsubsection{Object Detection}
We use a described object detector to find the set of domain- and task-relevant objects $\mathcal{O}$.
Described object detection \cite{describedod} unifies open-vocabulary object detection and phrase grounding, where the former detects all instances that match a text query class (e.g. cucumber), and the latter detects the single instance which matches the referring text phrase (e.g. the cucumber to the right of the bowl).
We need the former for detecting all objects that are domain-relevant, i.e. included in types $D_t$, and the latter for uniquely grounding the instances that are mentioned in the goal specification/instruction $G$.
Hence, the object detector is queried on image $I$ with types $D_t$ as class queries and phrases extracted from $G$ as phrase grounding queries.
The output consists of two sets of bounding boxes, one based on $D_t$, the other based on $G$.
We merge the detections by matching each box based on $G$ with one box based on $D_t$ by maximum Intersection over Union (IoU) to obtain the set of $N$ vertices $V$ for the spatial scene graph:
\begin{equation}
    \label{eq:vertices}
    V=\mathcal{O}=\{(t_i,n_i,b_i)\}_{i=1}^{N},
\end{equation} where each vertex has a type $t_i$, name $n_i$ and box $b_i$.
PDDL Objects can be written from this: $\{(n_i,t_i)\}_{i=1}^{N},$ e.g. \texttt{cucumber-vegetable} or \texttt{white\_bowl-container}.
\subsubsection{Spatial to Semantic Scene Graph} 
Given the vertices $V$ from object detection, we construct triplets of the form
$(v_i,r^s_{ij},v_j)$ to model a spatial relation between two vertices $v_i,v_j$.
Within the experiments of this paper, we simply calculate coordinate-wise difference $r^s_{ij} = b_i-b_j \in R^s \subseteq \mathds{R}^4$ for the spatial relation, but this could also respect more advanced geometric features as discussed in \Cref{sec:extendability}.

The goal is now to map this spatial scene graph to semantic scene graph $(V,E)$, with edges $E$ built from domain-relevant predicates $D_p$.
We achieve this through conditioning on the domain's set of types $D_t$ and predicate functions $D_p$, using $D_p$'s type constraint on its object arguments.
For example, for a binary predicate like ``in'', the first object argument can be of general ``object'' type while the second must be of ``container'' type.
We leverage this constraint for the scene graph and consider only triplets that may form a valid predicate.
For each predicate $p \in D_p$, the set of valid edges (triplets) is built for both an example (one-shot) and the test image, resulting in $E_p^{test},E_p^{ex}$.
Their corresponding spatial relations $R_p^{s,test},R_p^{s,ex}$ are considered features for classifying semantic relationships $D_{p}$.
Say a triplet $e=(v_i,p,v_j)$ evaluates $True$ iff. $p(v_i,v_j)$ holds.
Then each test triplet $e \in E_p^{test}$ gets classified by evaluating the predicate of its nearest neighbor (NN) triplet in $E_p^{ex}$.
NN is calculated from Euclidean distance in the spatial feature space shared by $R_p^{s,test},R_p^{s,ex}$.
Finally, all edges that are classified as $True$ are included in scene graph $\mathcal{G}=(V,E)$:
\begin{equation}
    \label{eq:edges}
    E=\bigcup_{p}\{e \in E_p^{test} \mid eval({NN_p}(e))\}.
\end{equation}
Initial state can be obtained from this through rewriting as
\begin{equation}
    \label{eq:sinit}
    S_{init}=\{p(v_i,v_j) \mid (v_i,p,v_j)\in E\},
\end{equation}
and PDDL \texttt{(:init} is written using names $n$ contained in $v$ as per \Cref{eq:vertices}, e.g. \texttt{in cucumber white\_bowl}.

\subsection{Goal Parser}
\label{subsec:goalgen}
A model that understands the instruction $G$ is employed as the goal parser.
Following the majority of the baselines, we focus on text instructions, so it is straightforward to use an LLM
Extendability to other goal modalities is discussed in \Cref{sec:extendability}.
The LLM is prompted to generate goal state $S_{goal}$ in PDDL given the natural language instruction $G$ and the domain's object and predicate types $D_t$ and $D_p$.
All object names used within the LLM-generated $S_{goal}$ are passed to the object detector and merged as per the paragraph before \Cref{eq:vertices} to ground initial and goal state on the same references. 
\section{EXPERIMENTS}
\subsection{Metrics}
We evaluate effectiveness of intermediate state grounding and final task planning.
For \textbf{state grounding}, we measure precision and recall of predicted subject-relation-object triplets that form the initial state.

For \textbf{task planning}, task planning success is given if the generated task plan transforms the ground truth initial state to the ground truth goal state.
In most cases, a correct plan can only be issued with a fully correct state, hence task planning success is the harder metric.
State grounding and task planning form our main results, presented in \Cref{tab:srate}.

Additionally, for checking the generated PDDL, we adopt the measure of problem file validity and plan validity 
from \cite{vilain} and present results in \Cref{tab:rprobrplan}.

\subsection{Dataset}
Our evaluation is based on the Problem Description Generation dataset ProDG \cite{vilain}. 
It provides three domains ($D$), each defined by a PDDL domain file specifying object types, lifted predicates and lifted actions.
In each domain there is a set of problems to solve.
Each problem comes with an image ($I$) as observation and a natural language instruction ($G$).
\begin{table}[H]

\centering
\caption{ProDG-v characteristics: Observed Predicates, number of derived predicates, \texttt{\#A}: number of skills in skillset, \texttt{mmA}: median minimum actions (that are required to complete the task), \texttt{m\#O}: median number of objects.}
\label{tab:prodgv}
\begin{tabular}{@{}l|l|c|c|c|c@{}}
\toprule
\textbf{Domain} & \textbf{Observed Pr.} & \textbf{\#Der.Pr.} & \textbf{\#A} & \textbf{mmA} & \textbf{m\#O} \\
\midrule

Cooking & carry,at,sliced & 3 & 3 & 9 & 8 \\
Blocksworld & on & 5 & 4 & 8 & 5 \\
Hanoi & on,onpeg,smaller & 2 & 1 & 47 & 8.5 \\
\bottomrule
\end{tabular}

\end{table}
\begin{table*}[htbp]
\centering
\caption{\colorbox{gray!20}{State grounding results} in Precision\textbar Recall of scene graph triplets. \colorbox{blue!10}{Task planning results} in success rate. \textit{OV} denotes open-vocabulary, \textit{S2S} denotes our spatial to semantic classifier. \textit{IR} lists type of intermediate representation, specifically \texttt{T} text, $\mathcal{G}$ scene graph, and \texttt{(:p} PDDL predicates.}
\label{tab:srate}
\begin{tabular}{@{}llp{1.5cm}p{1.75cm}lcclcccc@{}}
    \toprule
    \textbf{Framework} & \textbf{Shot} & & & & & &  \multicolumn{1}{c}{} & \multicolumn{1}{c}{\cellcolor{blue!10}\textbf{Planner}} & \cellcolor{blue!10}\textbf{C} & \cellcolor{blue!10}\textbf{B} & \cellcolor{blue!10}\textbf{H} \\
    \midrule
    \text{Direct Planning}  & 1 & & & & & & \multicolumn{1}{c}{}&\multicolumn{1}{l}{GPT-4o}  &  0.82&  0.11&   0.08  \\
    \midrule
     &  &\multicolumn{2}{l}{\cellcolor{gray!20} \textbf{State Grounding}} &  \cellcolor{gray!20} \textbf{IR} &  \cellcolor{gray!20} \textbf{[C]ooking} & \cellcolor{gray!20} \textbf{[B]locksw.} &  \multicolumn{1}{c}{\cellcolor{gray!20} \textbf{[H]anoi}} & \multicolumn{1}{c}{}  &  &  &   \\ \midrule
    $\text{VILA}$\cite{vila}& 0 & LMM & {GPT-4o} & \texttt{T} &-&-&-&GPT-4o & 0.00                &  0.10&   0.20  \\ 
    \multirow{2}{*}{$\text{NeuroGround}$\cite{neuroground}} &  \multirow{2}*{1}& \multirow{2}*{LMM} & \multirow{2}*{GPT-4o}  & \multirow{2}*{\texttt{T}} &\multirow{2}*{\textbf{1.00}\textbar0.92}&\multirow{2}*{0.74\textbar0.86}&\multirow{2}*{0.41\textbar0.50}& GPT-4o&\textbf{1.00}&  0.37&   0.08    \\
    &  & && &&&&  Solver& 0.88&  0.61&   0.08  \\
    $\text{VeriGraph}$\cite{verigraph}   & 1 & LMM &  GPT-4o &  $\mathcal{G}$ &0.92\textbar0.92&0.60\textbar0.85&0.39\textbar0.32& GPT-4  & -&  -&   -  \\   
    \midrule  
    ViLaIn\cite{vilain}      &1   & {VLM \textrightarrow LLM} &  {GDINO +BLIP2 \textrightarrow GPT-4} &  \texttt{(:p} &0.79\textbar0.72&0.49\textbar0.42&0.85\textbar0.72& {Solver}       & 0.00           &   0.03&  0.10\\ 
    \midrule
    $\text{Ours}$   &  1 & $\text{DC-SGG}_{VQA}$ & {GDINO \textrightarrow GPT-4o} &  $\mathcal{G}$\textrightarrow\texttt{(:p} &\textbf{1.00}\textbar\textbf{1.00}&0.71\textbar0.79&0.41\textbar0.36& Solver   &  \textbf{1.00}&  0.32&  0.24 \\ 
    $\text{Ours}$&0 & DC-SGG$_{OV}$ & RLIPv2 &  $\mathcal{G}$\textrightarrow\texttt{(:p} &0.93\textbar0.58&0.57\textbar0.73&0.70\textbar0.28&  {Solver}  &  0.00                &  0.00&   0.00\\
    $\text{Ours}$& 1 & $\text{DC-SGG}_{S2S}$ & GDINO\textrightarrow S2S  &  $\mathcal{G}$\textrightarrow\texttt{(:p} &\textbf{1.00}\textbar\textbf{1.00}&\textbf{0.98}\textbar\textbf{1.00}&\textbf{0.94}\textbar\textbf{0.91}& {Solver}   & 0.94&  \textbf{0.86}&   \textbf{0.65}\\ 
    \bottomrule
\end{tabular}

\end{table*}

To better suit the state grounding problem setting, we modify the original ProDG to create ProDG-v(isual).
This is necessary because the original ProDG dataset included (i) predicates that cannot be determined from observation, their truth value would need to be guessed based on few-shot examples, (ii) predicates that are always true across the dataset, diluting metrics.
Hence, in ProDG-v we distinguish between observed and derived predicates, where observed predicates can be \emph{v}isually perceived whereas derived predicates are merely a consequence of observed predicates or the consequence(effect) of an action.
Observed predicates can then be predicted from the image, while derived predicates are logically inferred based on domain definitions.

Summarized in \Cref{tab:prodgv}, the resulting ProDG-v dataset covers three domains with unique characteristics.
The Cooking domain resembles a real-world application where two robot arms must cooperate to slice and transport vegetables.
Blocksworld and Hanoi are both controlled environments, but they differ in complexity. Blocksworld is simpler, as it only requires determining the stack configuration of a few blocks and involves fewer actions. In contrast, Hanoi demands fine-grained distinctions between disk sizes, `on' vs. `on-peg' relationships, and, for some problems, requires a very large number of actions to rearrange the disks into the goal configuration.

\subsection{Implementation Details and Evaluation Settings}
We measure precision/recall of observed \textbf{predicates}.
Predicates that we consider are either unary or binary.
In our implementation, for binary predicates the logic from \Cref{eq:sinit} directly applies, for unary predicates we set $v_i=v_j$ for uniformity with the $(v_i,r,v_j)$ edge definition.
Instead of $b_i-b_j (\in \mathds{R}^4)$, the spatial feature is then created through differences from coordinate to coordinate within the box, resulting in a feature vector $\in \mathds{R}^6$.
For all one-shot experiments, to ensure there is a positive and negative reference, we sample a problem from ProDG-v that contains each predicate type occurring in the test image at least once and is not always true.

To achieve the described \textbf{object detector} functionality, we use models from the Grounding DINO family.
The class queries are passed to DINO-X \cite{dinox} and the goal phrases to Grounding DINO 1.0.
This is because DINO-X is stronger, but its API does not provide phrase grounding mode currently.
For fair comparison with the baseline \cite{vilain} that uses GroundingDINO for class queries, we upgrade its object detector to DINO-X, too.
For the LLM in the \textbf{goal parser} we use GPT-4o-mini which is sufficient for the task to convert the language instruction to PDDL format.
Except for methods that are designed as zero-shot task planners or unless otherwise stated, all experiments are conducted in the one-shot setting, where one input-output example is provided.
For the baselines, when we mention GPT-4o, we refer to gpt-4o-2024-11-20.
For the symbolic \textbf{planner}, we use Fast Downward \cite{fastdownward}, marked as ``Solver''.
As we are interested in evaluating the quality of state grounding, we do not allow re-prompting with error feedback from the planner to post-hoc alter state estimation.

\subsection{Comparison with Baselines}
\label{sec:lmmbaselines}
In \Cref{subsubsec:pddlbasedbaseline}, we compare against the baseline framework that generates a PDDL state with an LLM informed by VLMs.
In \Cref{subsubsec:llmasplannerbaseline}, we compare with frameworks that use an LMM as both state estimator and planner.

\subsubsection{PDDL State Generation}
\label{subsubsec:pddlbasedbaseline}
\begin{figure}[htpb]
    \centering
    \includegraphics[width=\linewidth]{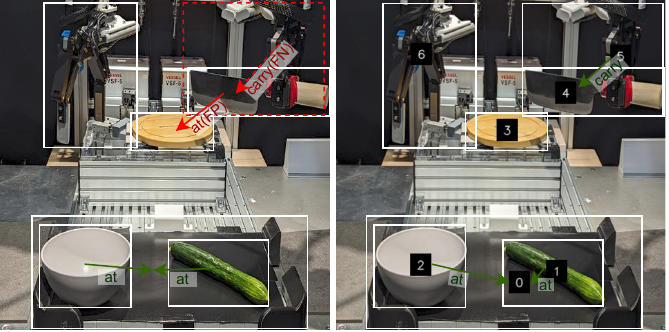}
    \caption{
    Left: ViLaIn \cite{vilain} detects false positive (FP) relation ``knife-at-board'' and misses (FN) relation ``right\_gripper-carry-knife''. Moreover, through its text-generative nature it can produce ungrounded hallucinations or forget already detected objects, here it forgets to include the ``right\_gripper'', dashed red box. Right: \emph{DC-SGG}$_{VQA}$ with SOM labels to uniquely refer to instances while querying LMM.}
    \label{fig:sgviz}
\end{figure}

With respect to the problem setting, \textbf{ViLaIn} \cite{vilain} is the most similar work to ours since their goal is to generate a PDDL problem consisting of objects, initial and goal state.
Their methodological approach is, however, very different.
It can be categorized under ``reasoning with LLM informed by perception data'', requiring a GroundingDINO object detector, a BLIP2-captioner, and a GPT-4 reasoner.
As shown in \Cref{tab:srate}(right), its success rates on task planning are $\leq 10\%$.
Since they use the same planner as ours, the issue lies in its poor state grounding ability (middle).
It suffers from three issues.
First, the captioner is not conditioned on task and domain, producing irrelevant captions that do not allow for subsequent inference of the relevant predicates.
Second, an LLM needs to write the state in PDDL, which can lead to violation of type rules or hallucinated or missing objects, documented in \Cref{fig:sgviz} and problem validity scores  $<1$ in \Cref{tab:rprobrplan}.
Third, their decision to generate the goal state after initial state leads to error propagation from initial to goal state.
All three problems are avoided by our state grounding framework design through (i) not relying on a captioner, (ii) writing initial state with DC-SGG instead of LLM, (iii) writing the goal state first.

\begin{table}[htbp]
    \centering
\caption{Ratio of valid (parsable) PDDL problems and plans. Generating PDDL problems with LLM may lead to syntax or type errors, while mapping from scene-graph to PDDL follows rules. On the other hand, the primary reason for the low number of valid plans is errors in state grounding, which cause the planning problem to have no solution.}
\label{tab:rprobrplan}
\begin{tabular}{@{}lcccccc@{}}
\toprule
\multirow{2}*{\textbf{Framework}} & \multicolumn{3}{c}{\textbf{Problems}} & \multicolumn{3}{c}{\textbf{Plans}} \\
                    & {C} & {B} & {H}  & {C} & {B} & {H} \\ \midrule
ViLaIn\cite{vilain}   &  0.94                &  0.88  & 0.94                  &   0.00 & 0.34 &  0.49   \\
Ours   &  1.00                &  1.00  & 1.00                    &   1.00 & 0.95  & 0.72     \\
\bottomrule
\end{tabular}
\end{table}

\subsubsection{LMM-As-Planner and State Estimator}
\label{subsubsec:llmasplannerbaseline}
We test frameworks that use LMMs for planning, starting with a direct approach, followed by approaches with an LMM that plans on its self-generated intermediate representation.

\textbf{Direct Planning} refers to the system of \Cref{fig:intro}a), where the task plan is directly predicted from image, domain, goal specification.
We provide the skill set and one example.
It is comparatively successful for the Cooking domain,
where the domain definition aligns with the LMM's common sense reasoning.
However, for Blocksworld, the direct planning attempt fails to respect the actual functionality of available skills, misusing skills in a way where preconditions are not fulfilled and effects do not align with the model's intention.
In Hanoi, long action sequences are often incomplete.

\textbf{VILA} \cite{vila}
ask the model to first list task-relevant objects.
Their reported prompt includes a rough description of constraints specific to their setups, so we need to adapt these sentences to our domain specification.
Since we assume domain specification in PDDL format, but \cite{vila} in natural language, we let an LLM generate a brief text description from the PDDL that matches the style of their prompt and verify that it is reasonable.
We can only measure task planning, not state grounding, because their intermediate output does not contain triplet information.
Results show that GPT-4o does not follow the domain-specific constraints for state grounding and planning, causing generation of invalid actions and leading to poor success rates $\leq20\%$ on all domains.
\textbf{NeuroGround} \cite{neuroground} also uses one LMM for both state grounding and planning, but additionally assists plan generation with a symbolic engine similar to our solver.
We evaluate their two variants LMM as planner vs. symbolic engine as planner.
We observe that the symbolic engine can avoid LMM planning mistakes in the Blocksworld domain ($0.61$ vs. $0.37$) where the LLM often misinterprets the functionality of the four actions stack, unstack, pickup, putdown.
However, at the same time parsing issues arise from the mismatch of state representation in weakly structured text modality and the highly structured language in which the parser and engine operate.
An effect of this is that the LMM-planning succeeds in the cooking domain while its generated state cannot be parsed for the engine so that the engine variant scores lower here ($1.00$ vs. $0.88$).

\begin{figure}[htbp]
    \centering
    \includegraphics[width=.9\linewidth]{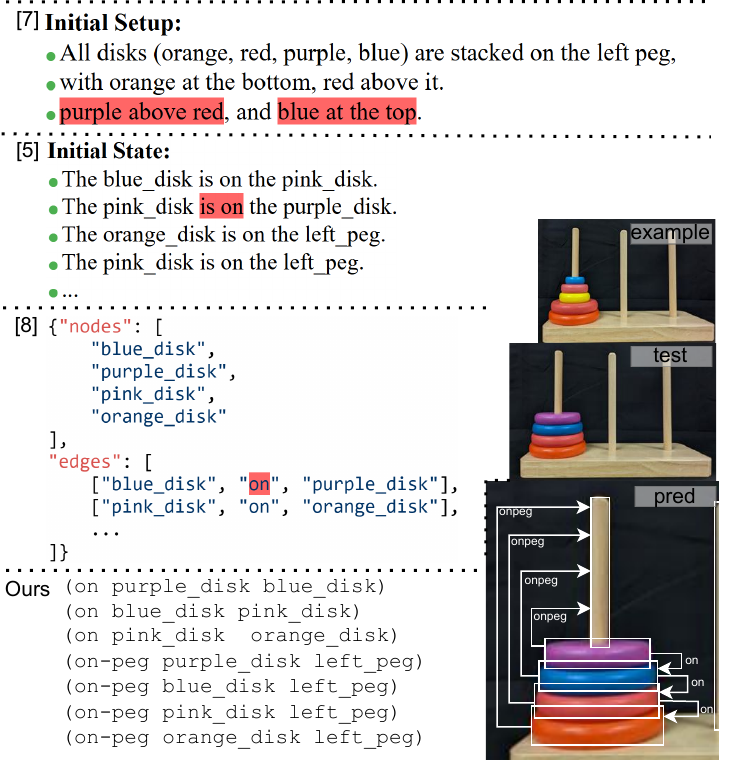}
    \caption{The methods \cite{vila, neuroground,verigraph} suffer from GPT-4o's mistakes in determining the initial state for the test image (wrong predictions highlighted in red). \cite{vila,neuroground} generate initial state as text representation, \cite{verigraph} as scene graph through coding JSON, ours as domain-conditioned scene graph (bottom right), corresponding to a PDDL init state (bottom left).}
    \label{fig:stategroundingerr}
\end{figure}

\textbf{VeriGraph} \cite{verigraph} operates, more similar to ours, strictly two-staged by first generating a scene graph and then planning on it.
However, their planning is implemented as prompting the LMM for graph editing, and they need multiple iterations to correct errors through feedback from an additional validator function.
As we do not allow these re-planning attempts, we only measure the state grounding accuracy.
As with other GPT-4o-based methods, state grounding is accurate for the Cooking domain, while for Hanoi it fails to extract the correct number, color, and position of disks and pegs, leading to errors similar to those seen in other models like \cite{vila,neuroground}.
We present an example in \Cref{fig:stategroundingerr}. 

In summary, LMM-based state grounding struggles with domain-specific, object-level scene understanding and larger numbers of objects and relations.
LMM-As-Planner struggles with adherence to domain constraints and longer plans.

\begin{table}[htbp]
   \centering
\caption{Grounded VQA experiments. Precision\textbar Recall on scene graph triplets.}
\label{tab:vqa}
\begin{tabular}{@{}llcccc@{}}
\toprule
\textbf{Framework} & \textbf{Model} & \textbf{Cooking} & \textbf{Blocksw.} & \textbf{Hanoi} \\ \midrule
\multirow{3}*{$\text{DC-SGG}_{VQA}$} & GPT-4o SOM  &  1.00\textbar1.00                &  0.71\textbar0.79                    &   0.41\textbar0.36                              \\
& MiniCPM-o SOM  &   0.92\textbar0.89  &  0.67\textbar0.73                &  0.43\textbar0.31                     \\
& BLIP3  &  0.83\textbar0.84                &  0.60\textbar0.75                    &   0.33\textbar0.37                              \\ 
\bottomrule
\end{tabular}
\end{table}
\begin{figure}[htbp]
    \centering
    \includegraphics[width=.8\linewidth]{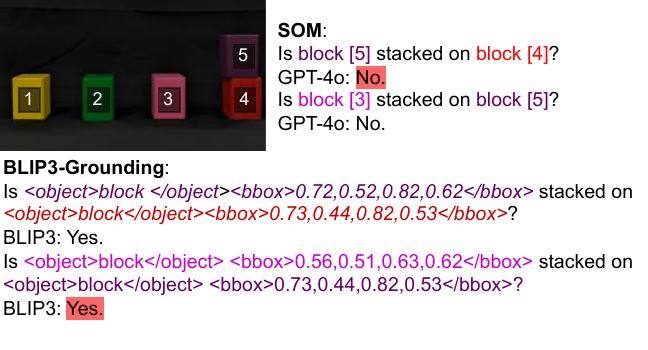}
    \caption{Experiment on replacing our predicate classification through visual question answering. Even when informed about locations through the object detector, LMMs surprisingly fail on elementary questions about spatial configuration, preventing them from being efficient scene parsers for state grounding. Image is from Blocksworld domain with mark overlays for SOM prompting.}
    \label{fig:blockswvqa}
\end{figure}

\subsection{Experiment on Predicate Classification through VQA}
\label{sec:vqaexperiment}
To provide more evidence that LMMs struggle with the required domain-specific, granular scene understanding, we show results for replacing our predicate classification through visual question answering (VQA) with an LMM, marked as \emph{DC-SGG}$_{VQA}$ in \Cref{tab:srate}.
Specifically, we ask a yes/no question for each possible grounded predicate.
The initial state is then the set of predicates where the LMM answered yes.
Since the questions are about specific instances that have been identified by the object detector, they must include references that can be uniquely associated with each instance.
Therefore we use set-of-marks (SOM) \cite{som} prompting for GPT-4o/MiniCPM-o-2\_6 \cite{minicpmo} and also try xGen-MM(BLIP3) \cite{blip3} which was trained with interleaved bounding boxes.
We show quantitative comparison in \Cref{tab:vqa}.
With an example in \Cref{fig:sgviz}, state grounding results are better compared to ViLaIn for Blocksworld and Cooking, but the approach is still ineffective for two reasons.
First, for predicates that take two object arguments, the set of grounded predicates grows quadratically w.r.t. objects.
This forbids joint reasoning over the whole predicate set.
Questions must be passed individually or batched, leading to the possibility of contradictory combinations like one disk being stacked on two pegs at the same time.
Second, as shown in \Cref{fig:blockswvqa}, there are still various obvious wrong answers, suggesting a lack of elementary object-level perception ability.
A series of other works \cite{spatialvlm,llavagrounding,spatialrgpt} towards models with improved spatial and grounded question answering may help to eventually resolve this issue.
Currently, however, with parameters below 8B, we found them to be inferior compared to \Cref{tab:vqa}.

\subsection{Experiment on Open-Vocabulary SGG}
In the above \Cref{sec:lmmbaselines} and \Cref{sec:vqaexperiment} we have observed that LMM-based methods face challenges as the number of objects and predicates in the scene increases.
Classic scene graph generation methods do not have this problem as they natively operate on a large set of object and relationship proposals.
Recent open-vocabulary scene graph generators like RLIPv2 \cite{rlipv2} or Scene Graph ViT \cite{sgvit} are built similarly to open-vocabulary object detectors with additional relationship support, allowing to match object/relationship proposals with text queries.
Open-vocab SGG would therefore constitute a more elegant solution to unify our scene parser in a single model.
In \emph{DC-SGG}$_{OV}$ we experiment with RLIPv2 \cite{rlipv2} ParSeDA, Swin-L as zero-shot SGG as we identify no simple way to feed the one-shot example.
The \texttt{verb} text queries are the predicate names within the domain specification and the \texttt{object} text queries are the same as for our object detector.
Entity predictions are subjected to non-maximum suppression, and we set the score threshold for including a triplet score in the final predictions based on the empirically optimal value for each domain.
Same as for our predicate classifier, we use the domain information to consider only triplets that may form a valid grounded predicate.
State grounding results in \Cref{tab:srate} show that in principle the solution is possible.
However, no scene graph is entirely correct, which would be a necessity for plan success, leading to the reported zero sucess rates.
RLIPv2 struggles with the long-tail object types like the chopping knife in the Cooking domain or the peg in the Hanoi domain.
Moreover, the meaning of predicate names might be ambiguous, e.g. one disk ``on'' another disk could mean directly on top or merely somewhere above in the stack.
We see efforts \cite{fsod} to develop an interface for aligning object-centric open-vocab models with the target concept as a promising direction to tackle this issue.

\section{Discussion} 
\subsection{Extendability}
\label{sec:extendability}
We briefly explore the potential for extended variants.\\ 
\textbf{3D.}
Our approach classifies predicates based on spatial features.
In the dataset used, relevant relationships are distinguishable in 2D.
However, in other environments, it may be necessary to incorporate 3D information to classify predicates.
In this case, one can build spatial features from point clouds using a depth sensor, similar to the approach described in \cite{spatialvlm}.
\textbf{Masks.}
Alternatively, in environments where fine-grained spatial information is more important, features could be constructed using segmentation masks, such as those produced by GroundedSAM, rather than relying solely on DINO-X bounding boxes.
\textbf{Goal Modality.}
In this paper we evaluated under instructions in text form, but our framework  also supports goal images.
The LLM-based goal parser is then replaced through a second scene parser from \Cref{subsec:sceneparser} that determines the goal state through generating a scene graph of the goal image.
Besides natural language and goal images, logical goal conditions as in \cite{geometricsymbolic} can be directly provided, essentially skipping the goal parser.
\textbf{Closed Loop.}
Experiments in this paper were limited to an open-loop setting, but our framework can also run in a loop - more resource efficient than LMM approaches.
Only the scene parser runs again to extract a new estimate of object set and initial state.
The goal remains constant throughout execution, so the goal parser LLM only needs to be run once at the start.

\subsection{Limitations and Future Work}
The evaluations in this paper were performed based on the dataset from the baseline \cite{vilain}, covering hand-crafted manipulation scenes.
We acknowledge the quantitative results shown have limited implications on other task families and therefore in the next step, we want to evaluate in simulation, where we can (i) validate that the method works together with existing low-level skills and (ii) generate a dataset controlling for configurations to allow for more systematic analysis.

\section{CONCLUSION}
In this paper, we present a state grounding framework for robotic task planning that parses scene observation and instruction into PDDL to transform the planning problem into an algorithmically solvable problem.
For parsing the scene observation into the required plannable states, we introduce domain-conditioned scene graphs as a structure mappable to PDDL.
These scene graphs are generated in a more classical way with object detection and predicate classification.
Comparison with approaches relying on large multimodal models reveals that they still face major weaknesses in domain-specific visual perception and planning under constraints.
We hope our work can aid future research in addressing these shortcomings in state-grounded task planning.






\renewcommand*{\bibfont}{\footnotesize}
\printbibliography

\end{document}